\documentclass{article}

\usepackage[final, nonatbib]{nips_2018_ml4h}
\usepackage[utf8]{inputenc}
\usepackage[T1]{fontenc}
\usepackage{hyperref}
\usepackage{url}
\usepackage{booktabs}
\usepackage{amsfonts}
\usepackage{amssymb}
\usepackage[sort,numbers]{natbib}
\usepackage{nicefrac}
\usepackage{microtype}
\usepackage{multirow}
\usepackage{subcaption}
\usepackage{todonotes}

\title{Unsupervised Multimodal Representation Learning across Medical Images and Reports}

\author{
    Tzu-Ming Harry Hsu,
    Wei-Hung Weng,
    Willie Boag,
    Matthew McDermott,
    Peter Szolovits
    \\
    MIT CSAIL \\
    \texttt{\{stmharry,ckbjimmy,wboag,mmd,psz\}@mit.edu} 
}

\begin{document}

\maketitle
\begin{abstract}
Joint embeddings between medical imaging modalities and associated radiology reports have the potential to offer significant benefits to the clinical community, ranging from cross-domain retrieval to conditional generation of reports to the broader goals of multimodal representation learning. 
In this work, we establish baseline joint embedding results measured via both local and global retrieval methods on the soon to be released MIMIC-CXR dataset consisting of both chest X-ray images and the associated radiology reports. 
We examine both supervised and unsupervised methods on this task and show that for document retrieval tasks with the learned representations, only a limited amount of supervision is needed to yield results comparable to those of fully-supervised methods.

\end{abstract}
\section{Introduction}

Medical imaging is one of the most compelling domains for the immediate application of artificial intelligence tools. 
Recent years have seen not only tremendous academic advancements \cite{esteva2017dermatologist,gulshan2016development,rajpurkar2017chexnet} but additionally a breadth of applied tools \cite{marr2017first,Walter2018DensitasDensity,Lagasse2018FDANews,EnvoyAI2017EnvoyAIPartners}. 

There has been some emerging attention on joint processing of medical images and radiological free-text reports. \cite{wang2018tienet} used the public NIH Chest X-ray 14 dataset~\citep{wang2017chestx} linked with the non-public associated reports to both improve disease classification performance and for automatic report generation. \cite{gale2018producing} attempted to generate radiology reports while \cite{shin2016learning} generated disease/location/severity annotations.
\cite{liu2018learning} generated notes, including radiology reports for the Medical Information Mart for Intensive Care (MIMIC) dataset using non-image modalities such as demographics, previous notes, labs, and medications.
These works used annotations from either machines~\citep{wang2017chestx} or humans. 
However, with a huge influx of imaging data beyond human capacity, parallel records from both imaging and text are not always readily available.
We thus would like to bring up the question of whether we can take advantage of unannotated but massive
imaging datasets and learn from the underlying distribution of these images. 

One natural area that remains unexplored is representation learning across images and reports. 
The idea of representation learning in a joint embedding space can be realized in multiple ways. 
Some~\citep{pan2011domain,chen2016transfer} explored statistical and metrical relevance across domains, and some~\citep{ganin2016domain} realized it as an adversarially determined domain-agnostic latent space.
\cite{shen2017style,mor2018universal} both used a the latent space for style transfer, in language sentiment and music style, respectively.
\cite{reed2016learning} learned joint spaces of images and their captions, which \cite{reed2016generative} later used for caption-driven image generation. \cite{conneau2017word} and \cite{grave2018unsupervised} also used similar ideas to perform both supervised and unsupervised word-to-word translation tasks.
\citep{chung2018unsupervised} further aligned cross-modal embeddings through semantics in speech and text for spoken word classification and translation tasks.

A recent dataset, MIMIC- Chest X-ray\footnote{Soon to be publicly released.} (MIMIC-CXR), carries paired records of X-ray images and radiology reports, and the imaging modality has been explored in~\cite{rubin2018large}. 
In this work, we explore both the text and image modalities with joint embedding spaces under a spectrum of supervised and unsupervised methods.
In particular, we  make the following contributions:

\begin{enumerate}
\item We establish baseline results and evaluation methods for jointly embedding radiological images and reports via retrieval and distance metrics.
\item We profile the impact of supervision level on the quality of representation learning in joint embedding spaces.
\item We characterize the influence of using different sections from the report on representation learning.
\end{enumerate}

\section{Methodology}

\subsection{Data}
All experiments in this work used the MIMIC-CXR dataset. MIMIC-CXR consists of 473,057 chest X-ray images and 206,563 reports from 63,478 patients. 
Of these images, 240,780 are of anteroposterior (AP) views, which we focus on in this work. 
Further, we eliminate all duplicated radiograph images with adjusted brightness or contrast\footnote{Commonly produced for clinical needs}, leaving a total of 95,242/87,353 images/reports, which we subdivide into a train set of 75,147/69,171 and a test set of 19,825/18,182 images/reports, with no overlap of patients between the two.
Radiological reports are parsed into sections and we use either the \emph{impression} or the \emph{findings} sections.

For evaluation, we aggregate a list of unique International Classification of Diseases (ICD-9) codes from all patient admissions and ask a clinician to pick out a subset of codes that are related to thoracic diseases. Records with ICD-9 codes in the subset are then extracted, including 3,549 images from 380 patients. This population serves as a disease-related evaluation for retrieval algorithms. Note that this disease information is never provided during training in any setting.

\subsection{Methods}

\begin{figure}[!t]
    \centering
    \includegraphics[width=0.8\textwidth]{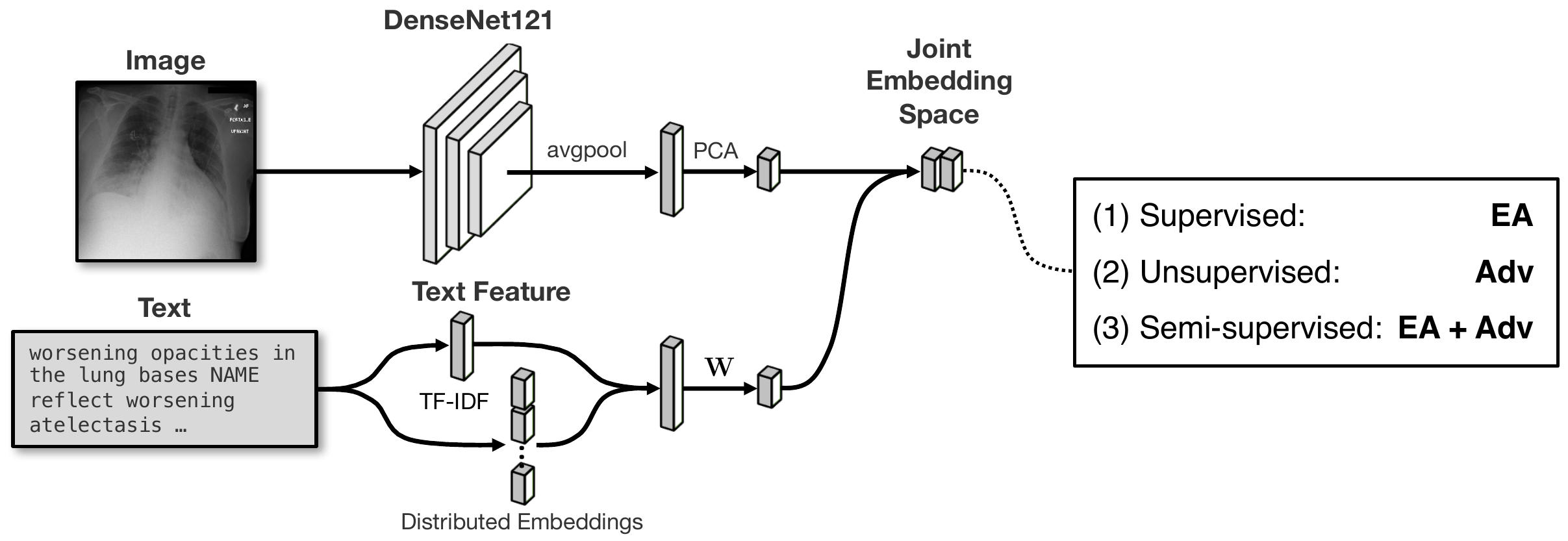}
    \caption{The overall experimental pipeline. EA: embedding alignment; Adv: adversarial training.}
    \label{fig:pipeline}
\end{figure}

Our overall experimental flow follows Figure~\ref{fig:pipeline}. 
Notes are featurized via (1) term frequency-inverse document frequency (TF-IDF) over bi-grams, (2) pre-trained GloVe \emph{word} embeddings~\citep{pennington2014glove} averaged across the selected section of the report, (3) \emph{sentence} embeddings, or (4) \emph{paragraph} embeddings. In (3) and (4), we first perform sentence/paragraph splitting, and then fine-tune a deep averaging network (DAN) encoder~\citep{bird2004nltk,cer2018universal,iyyer2015deep} with the corpus. Embeddings are finally averaged across sentences/paragraphs.
The DAN encoder is pretrained on a variety of data sources and tasks and fine-tuned on the context of report sections.

Images are resized to 256$\times$256, then featurized to the last bottleneck layer of a pretrained DenseNet-121 model~\citep{rajpurkar2017chexnet}. 
PCA is applied onto the 1024-dimension raw image features to obtain 64-dimension features.\footnote{96.9\% variance explained}
Text features are projected into the 64-dimension image feature space. We use several methods regarding different objectives.

\paragraph{Embedding Alignment (EA)} 
Here, we find a linear transformation between two sets of matched points $\mathbf{X} \in \mathbb{R}^{d_X \times n}$ and $\mathbf{Y} \in \mathbb{R}^{d_Y \times n}$ by minimizing $\mathcal{L}_{\mathrm{EA}} \left(\mathbf{X}, \mathbf{Y}\right) = \left\|\mathbf{W}^\top\mathbf{X} - \mathbf{Y}\right\|_F^2$. 

\paragraph{Adversarial Domain Adaption (Adv)}
Adversarial training pits a \emph{discriminator}, $D$, implemented as a 2-layer (hidden size 256) neural network using scaled exponential linear units (SELUs) \citep{klambauer2017self}, against a projection matrix $\mathbf{W}$, as the \emph{generator}. $D$ is trained to classify points in the joint space according to source modality, and $\mathbf W$ is trained adversarially to fool $D$. Alternatively, $D$ minimizes $\mathcal{L}_{\mathrm{Adv}}^D \left(\mathbf{X}, \mathbf{Y}\right) = \mathbb{E}_{\left(\mathbf{x}, \mathbf{y}\right) \sim p\left(\mathbf{X}, \mathbf{Y}\right)} \left[-\log D\left(\mathbf{W}^\top\mathbf{x}\right) - \log\left(1 - D\left(\mathbf{y}\right)\right)\right]$ when $\mathbf{W}$ minimizes $\mathcal{L}_{\mathrm{Adv}}^{\mathbf{W}} \left(\mathbf{X}, \mathbf{Y}\right) = \mathbb{E}_{\left(\mathbf{x}, \mathbf{y}\right) \sim p\left(\mathbf{X}, \mathbf{Y}\right)} \left[-\log\left(1 - D\left(\mathbf{W}^\top\mathbf{x}\right)\right)\right]$. 

\paragraph{Procrustes Refinement (Adv + Proc)}
On top of adversarial training, we also use an unsupervised Procrustes induced refinement as in~\cite{conneau2017word}.

\paragraph{Semi-Supervised}
We also assess how much supervision is necessary to ensure strong performance on these modalities by randomly subsampling our data into supervised and unsupervised samples.
We then combine the \emph{embedding alignment} objective and \emph{adversarial training} objective functions as $\mathcal{L} = \mathcal{L}_{\mathrm{EA}} \left(\mathbf{X}, \mathbf{Y}\right) + \lambda \mathcal{L}_{\mathrm{Adv}} \left(\mathbf{X}, \mathbf{Y}\right) $ and train simultaneously as we vary the fraction trained. Preliminary experiments suggests $\lambda = 0.1$.

\paragraph{Orthogonal Regularization}
\cite{smith2017offline,conneau2017word,xing2015normalized} all showed that imposing orthonormality on linear projections leads to better performance and stability in training . However,~\cite{brock2018large} suggested orthogonality (i.e., not constraining the norms) can perform better as a regularization. Thus on top of the objectives, we add $\mathcal{R}_{\mathrm{ortho}} = \beta \left\|\mathbf{W}^\top\mathbf{W} \odot \left(\mathbf{e}\mathbf{e}^\top - \mathbf{I}\right)\right\|_F^2$, where $\odot$ denotes element-wise product and $\mathbf{e}$ denotes a column vector of all ones. Scanning through a range shows $\beta = 0.01$ yields good performance.

\subsection{Evaluation}
We evaluate via cross domain retrieval in the test set $Q$: querying in the joint embedding space for closest neighboring images using a report, $\mathsf{T}\rightarrow\mathsf{I}$, or vice-versa, $\mathsf{I}\rightarrow\mathsf{T}$.
For direct pairings, we compute the \emph{cosine similarity}, and $\mathsf{MRR} = \frac{1}{\left|Q\right|} \sum_{q \in Q} \frac{1}{\mathsf{rank}_q}$ where $\mathsf{rank}_q$ is the rank of the first true pair for $q$ (e.g., the first paired image or text corresponding to the query $q$) in the retrieval list.
For thoracic disease induced pairings, we first define the relevance $\mathsf{rel}_{pq} \in \left[ 0, 1 \right]$ between two entries $p$ and $q$ as the intersection-over-union of their respective set of ICD-9 codes. 
Then we calculate the normalized discounted cumulative gain~\citep{jarvelin2002cumulated} $\mathsf{nDCG@k} = \frac{1}{\left|Q\right|} \sum_{q \in Q} \frac{1}{\mathsf{IDCG}_q} \sum_{p=1}^k \frac{2^{\mathsf{rel}_{pq}} - 1}{\log_2 \left(p + 1\right)}$, where $\mathsf{IDCG}_q$ denotes the ideal $\mathsf{DCG}$ value for $q$ using a perfect retrieval algorithm.
All experiments are repeated with random initial seeds for at least 5 times. Means and 95\% confidence intervals are reported in the following section.

\begin{table}[!t]
    \centering\scriptsize
    \setlength{\tabcolsep}{2pt}
    \begin{tabular}{*{11}{c}}
        \toprule
        \multirow{2}[2]{*}{Text Feature}
        & \multirow{2}[2]{*}{Method}
        & \multirow{2}[2]{*}{Similarity}
        & \multicolumn{2}{c}{$\mathsf{MRR}(\times 10^{-3})$}
        & \multicolumn{2}{c}{$\mathsf{nDCG@1}$}
        & \multicolumn{2}{c}{$\mathsf{nDCG@10}$}
        & \multicolumn{2}{c}{$\mathsf{nDCG@100}$}
        \\ \cmidrule(lr){4-5} \cmidrule(lr){6-7} \cmidrule(lr){8-9} \cmidrule(lr){10-11}
        &&
        & $\mathsf{T}\rightarrow\mathsf{I}$ & $\mathsf{I}\rightarrow\mathsf{T}$
        & $\mathsf{T}\rightarrow\mathsf{I}$ & $\mathsf{I}\rightarrow\mathsf{T}$
        & $\mathsf{T}\rightarrow\mathsf{I}$ & $\mathsf{I}\rightarrow\mathsf{T}$
        & $\mathsf{T}\rightarrow\mathsf{I}$ & $\mathsf{I}\rightarrow\mathsf{T}$
        \\ \midrule
        & \it chance
        &
        & $0.50$ & $0.50$
        & $0.103$ & $0.103$
        & $0.103$ & $0.103$
        & $0.103$ & $0.103$
        \\ \midrule
        bi-gram & EA & $\mathbf{0.613}_{.000}$ & $\mathbf{7.33}_{.04}$ & $\mathbf{11.65}_{.07}$ & $\mathbf{0.147}_{.001}$ & $\mathbf{0.162}_{.001}$ & $\mathbf{0.148}_{.000}$ & $\mathbf{0.159}_{.000}$ & $\mathbf{0.225}_{.000}$ & $\mathbf{0.231}_{.000}$ \\
word & EA & $0.542_{.000}$ & $2.00_{.01}$ & $4.52_{.02}$ & $0.096_{.002}$ & $0.128_{.001}$ & $0.116_{.000}$ & $0.130_{.000}$ & $0.202_{.000}$ & $0.205_{.000}$ \\
sentence & EA & $0.465_{.000}$ & $1.08_{.00}$ & $2.74_{.02}$ & $0.073_{.001}$ & $0.101_{.000}$ & $0.100_{.000}$ & $0.111_{.000}$ & $0.189_{.000}$ & $0.177_{.000}$ \\
paragraph & EA & $0.505_{.000}$ & $1.57_{.01}$ & $2.53_{.01}$ & $0.082_{.001}$ & $0.134_{.000}$ & $0.107_{.000}$ & $0.124_{.000}$ & $0.195_{.000}$ & $0.196_{.000}$ \\
\midrule 
bi-gram & Adv & $0.218_{.073}$ & $0.77_{.23}$ & $0.85_{.33}$ & $0.095_{.006}$ & $0.090_{.003}$ & $0.101_{.004}$ & $0.098_{.003}$ & $0.171_{.005}$ & $0.166_{.004}$ \\
bi-gram & Adv + Proc & $0.221_{.074}$ & $0.77_{.24}$ & $0.87_{.32}$ & $0.094_{.006}$ & $0.091_{.004}$ & $0.102_{.004}$ & $0.099_{.002}$ & $0.171_{.005}$ & $0.166_{.004}$ \\
word & Adv & $0.268_{.016}$ & $0.65_{.12}$ & $0.54_{.12}$ & $0.096_{.006}$ & $0.091_{.003}$ & $0.105_{.004}$ & $0.099_{.003}$ & $0.176_{.003}$ & $0.165_{.004}$ \\
word & Adv + Proc & $\mathbf{0.269}_{.013}$ & $0.64_{.11}$ & $0.57_{.07}$ & $\mathbf{0.098}_{.006}$ & $0.092_{.002}$ & $\mathbf{0.107}_{.005}$ & $0.099_{.003}$ & $\mathbf{0.179}_{.003}$ & $0.165_{.004}$ \\
sentence & Adv & $0.265_{.010}$ & $0.64_{.08}$ & $\mathbf{1.07}_{.24}$ & $0.095_{.007}$ & $0.094_{.002}$ & $0.103_{.006}$ & $0.100_{.001}$ & $0.176_{.006}$ & $0.167_{.001}$ \\
sentence & Adv + Proc & $0.266_{.012}$ & $0.68_{.10}$ & $1.07_{.21}$ & $0.096_{.005}$ & $0.094_{.004}$ & $0.105_{.006}$ & $0.100_{.002}$ & $0.178_{.005}$ & $0.166_{.002}$ \\
paragraph & Adv & $0.045_{.136}$ & $0.69_{.03}$ & $0.70_{.04}$ & $0.062_{.025}$ & $\mathbf{0.123}_{.029}$ & $0.082_{.015}$ & $\mathbf{0.118}_{.017}$ & $0.163_{.013}$ & $\mathbf{0.169}_{.003}$ \\
paragraph & Adv + Proc & $0.225_{.061}$ & $\mathbf{1.15}_{.60}$ & $0.77_{.21}$ & $0.093_{.057}$ & $0.092_{.011}$ & $0.090_{.034}$ & $0.103_{.008}$ & $0.163_{.023}$ & $0.166_{.005}$ \\

        \bottomrule
    \end{tabular}
    
    \vspace{10pt}
    \caption{Comparison among supervised (upper) and unsupervised (lower) methods. Subscripts show the half width of $95\%$ confidence intervals. \textbf{Bold} denotes the best performance in each group. \emph{Chance} is the expected value if we randomly yield retrievals. Higher is better for all metrics.}
    \label{table:retrieval}
    \vspace{-10pt}
\end{table}
\section{Results}

\paragraph{Retrieval with/without Supervision}
Table~\ref{table:retrieval} compares four types of text features and supervised/unsupervised methods. We find that unsupervised methods can achieve comparable results on disease-related retrieval tasks on a large scale ($\mathsf{nDCG@100}$) without the need for labeling the chest X-ray images. 
Experiments show uni-, bi-, and tri-grams yield very similar results and we only include bi-gram in the table. 
Additionally, we find that the high-level \emph{sentence} and \emph{paragraph} embeddings approach underperformed the bi-gram text representation. 
Although having generalizability~\citep{cer2018universal}, sentence and paragraph embeddings learned from the supervised multi-task pre-trained model may not be able to represent the domain-specific radiological reports well due to the lack of medical domain tasks in the pre-training process. 
Unsupervised procrustes refinement is occasionally, but not universally helpful.
Note that $\mathsf{MRR}$ is comparatively small since reports are in general highly similar for radiographs with the same disease types. 

\begin{figure}[!h]
    \vspace{-10pt}
    \centering
    \begin{subfigure}[t]{0.25\linewidth}
        \includegraphics[width=1.05\textwidth]{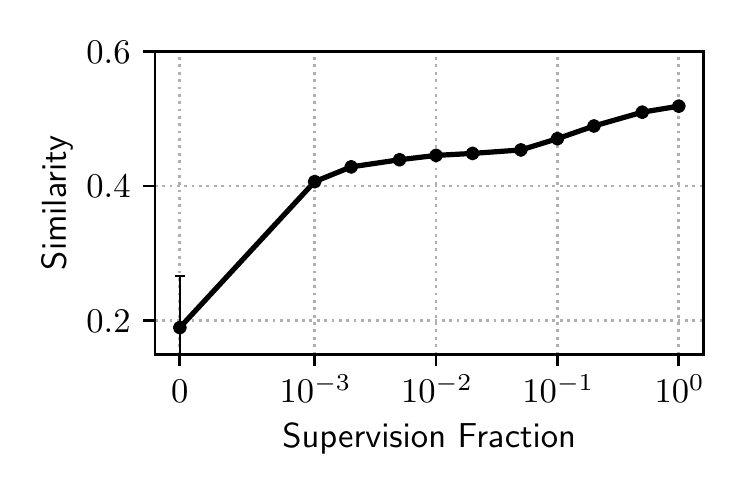}
    \end{subfigure}%
    \begin{subfigure}[t]{0.25\linewidth}
        \includegraphics[width=1.05\textwidth]{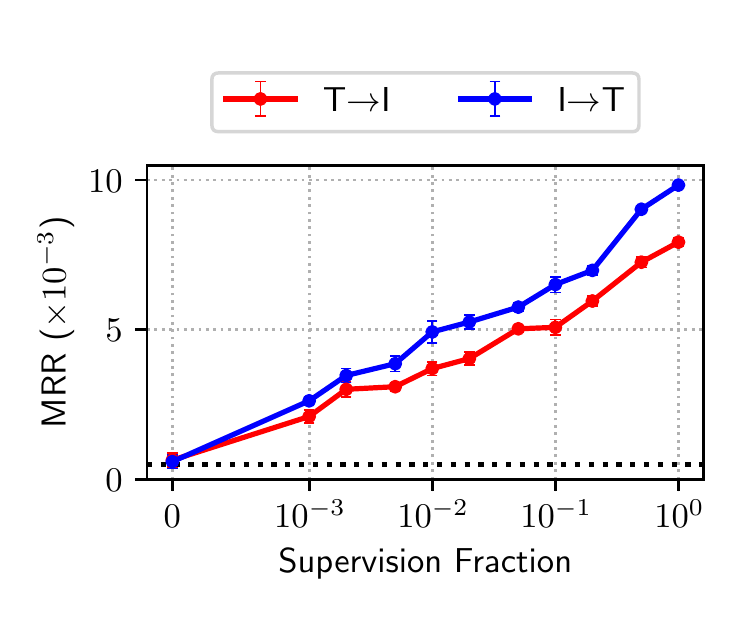}
    \end{subfigure}%
    \begin{subfigure}[t]{0.25\linewidth}
        \includegraphics[width=1.05\textwidth]{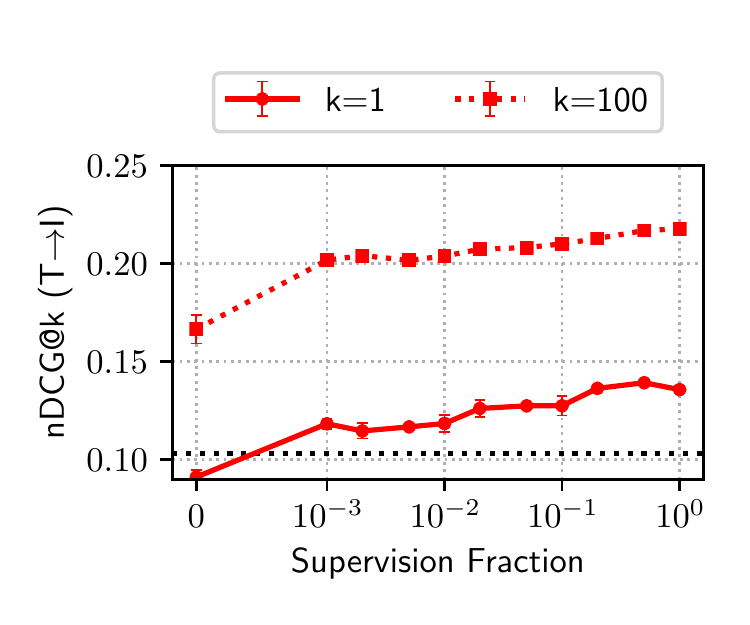}
    \end{subfigure}%
    \begin{subfigure}[t]{0.25\linewidth}
        \includegraphics[width=1.05\textwidth]{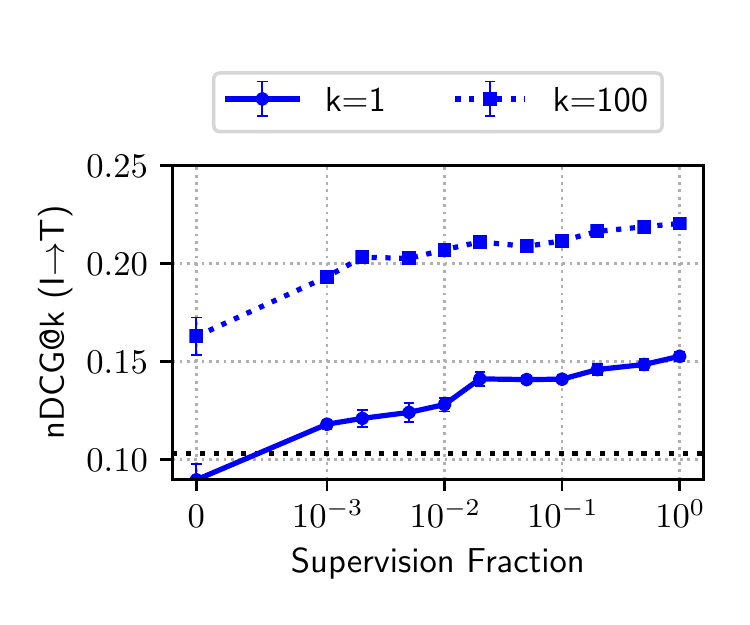}
    \end{subfigure}%
    \vspace{-10pt}
    \caption{Performance measures of retrieval tasks at $k$ retrieved items as a function of the supervision fraction. Higher is better. Note the $x$-axis is in log scale. Unsupervised is on the left, increasingly supervised to the right. Dashed lines indicate the performance by chance. Vertical bars indicate the 95\% confidence interval, and some are too narrow to be visible.}
    \label{fig:supervision}
\end{figure}

\paragraph{The Impact of Supervision Fraction}
We define the \emph{supervision fraction} as the fraction of pairing information provided in the training set. 
Note the ICD-9 codes are not provided for training even in the fully supervised setting.
Figure~\ref{fig:supervision} shows our evaluation metrics for models trained using bi-gram text features and the semi-supervised learning objective for various supervision fractions. 
A minimal supervision as low as 0.1\% provided can drastically improve the alignment quality, especially in terms of cosine similarity and $\mathsf{nDCG}$. 
More annotations further improve the performance measures, but one would almost require exponentially many data points in exchange for a linear increase.
That implies the possibility of concatenating a well-annotated dataset and a large but unannotated dataset for a substantial performance boost.

\begin{figure}[!h]
    \vspace{-10pt}
    \centering
    \includegraphics[width=0.9\textwidth]{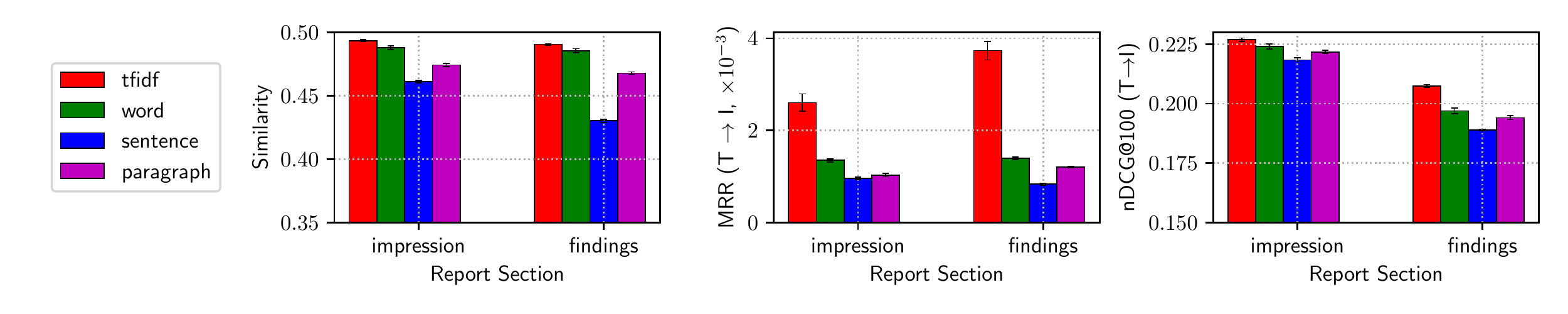}
    \vspace{-10pt}
    \caption{Different metrics for retrieval on either the \emph{impression} or \emph{findings} section using four types of features. 95\% confidence intervals are indicated on the bars.}
    \label{fig:section}
\end{figure}

\paragraph{Using Different Sections of the Report}
We investigate the effectiveness of using different sections for the embedding alignment task. All models in Figure~\ref{fig:section} run with a supervision fraction of 1\%.
The models trained on the \emph{findings} section outperformed the models trained on the \emph{impression} section using cosine similarity and $\mathsf{MRR}$. 
This makes sense from a clinical perspective since the radiologists usually only describe image patterns in the \emph{findings} section and thus they would be aligned well. 
On the other hand, they make radiological-clinical integrated interpretations in the \emph{impression} section, which means that the both the image-uncorrelated clinical history and findings were mentioned in the \emph{impression} section.
Since $\mathsf{nDCG}$ is calculated using ICD-9 codes, which carry disease-related information, it naturally aligns with the purpose of writing an \emph{impression} section.
This may explain why the models trained on \emph{impression} section worked better for $\mathsf{nDCG}$. 


\section{Conclusion}
MIMIC-CXR will soon be the largest publicly available imaging dataset consisting of both medical images and paired radiological reports, promising myriad applications that can make use of both modalities together. We establish baseline results using supervised and unsupervised joint embedding methods along with local (direct pairs) and global (ICD-9 code groupings) retrieval evaluation metrics. Results show a possibility of incorporating more unsupervised data into training for minimal-effort performance increase. A further study of joint embeddings between these modalities may enable significant applications, such as text/image generation or the incorporation of other EMR modalities.

\bibliographystyle{plain}
\bibliography{references,references_mendeley}
 
\end{document}